\documentclass{article}

\usepackage{microtype}
\usepackage{graphicx}
\usepackage{subcaption}
\usepackage{booktabs} %
\usepackage{multirow}

\usepackage{hyperref}

\usepackage[preprint]{icml2025}

\usepackage{amsmath}
\usepackage{amssymb}
\usepackage{mathtools}
\usepackage{amsthm}

\usepackage[capitalize,noabbrev]{cleveref}

\theoremstyle{plain}
\newtheorem{theorem}{Theorem}[section]

\theoremstyle{definition}

\theoremstyle{remark}

\definecolor{myred}{RGB}{251, 214, 211}
\definecolor{myblue}{RGB}{47, 110, 186}
\definecolor[named]{ACMPurple}{cmyk}{0.55,1,0,0.15}
\hypersetup{
  colorlinks=true,
  citecolor=ACMPurple
}

\usepackage[textsize=tiny]{todonotes}

\icmltitlerunning{Iterative Deepening Sampling as Efficient Test-Time Scaling}

\begin{document}

\twocolumn[
\icmltitle{Iterative Deepening Sampling as Efficient Test-Time Scaling}

\begin{icmlauthorlist}
\icmlauthor{Weizhe Chen}{usc}
\icmlauthor{Sven Koenig}{uci}
\icmlauthor{Bistra Dilkina}{usc}
\end{icmlauthorlist}

\icmlaffiliation{usc}{Thomas Lord Department of Computer Science, University of Southern California, United States}
\icmlaffiliation{uci}{Department of Computer Science, University of California, Irvine, United States}

\icmlcorrespondingauthor{Weizhe Chen}{weizhech@usc.edu}
\icmlcorrespondingauthor{Sven Koenig}{sven.koenig@uci.edu}
\icmlcorrespondingauthor{Bistra Dilkina}{dilkina@usc.edu}

\icmlkeywords{Machine Learning, ICML, LLM}

\vskip 0.3in
]

\printAffiliationsAndNotice{} %

\begin{abstract}
Recent reasoning models, such as OpenAI’s O1 series, have demonstrated exceptional performance on complex reasoning tasks and revealed new test-time scaling laws. Inspired by this, many people have been studying how to train models to achieve effective self-evaluation and self-correction to further enable the scaling paradigm. However, less studied is how to efficiently scale test-time compute from a fixed model, and this remains a challenge. In this paper, we address this challenge by focusing on enhancing the quality of self-reflection data generation for complex problem-solving at test time, which can also subsequently improve the training of next-generation large language models (LLMs). Specifically, we explore how systematically triggering a model's self-correction mechanisms can improve performance on challenging reasoning tasks. To this end, we propose a novel iterative deepening sampling algorithm framework designed to enhance self-correction and generate higher-quality samples. Through extensive experiments on Math500 and AIME benchmarks, we demonstrate that our method achieves a higher success rate on difficult tasks and provide detailed ablation studies to analyze its effectiveness across diverse settings. 
\end{abstract}

\section{Introduction}

Since ChatGPT, large language models (LLMs) have been a rapidly evolving domain that tries to solve problems beyond traditional language tasks like summarization or question answering \cite{chen2023teaching, yao2022react, chen2024alphamathzeroprocesssupervision, chen2024solving}. Significantly, the newly released OpenAI O1 has introduced its new paradigm of test-time scaling, which shows strong capability in complex problem-solving through its detailed reasoning steps before outputting the final answer \cite{jaech2024openai}. Since then, many researchers have studied how to replicate success from an open-source perspective and how to train models that are even better at efficiently solving problems that still remain unsolvable by the current LLMs \cite{huang2024o1, zeng2024scaling, deepseekai2025deepseekr1incentivizingreasoningcapability}. One key finding is that through reinforcement learning itself, LLM can spontaneously learn to self-evaluate and self-correct from time to time. However, there is no clear conclusion on whether self-evaluation is triggered often enough.

On the other hand, while training is the primary focus recently, it remains uncertain whether one could more efficiently scale test-time compute from a fixed model without additional changes in training or fine-tuning. 
Moreover, a more efficient sampling algorithm used at test-time can not only enhance inference-time efficiency but also facilitate the generation of high-quality synthetic data, which can be leveraged to train the next generation of models and evaluate their performance \cite{guan2025rstar}.

\begin{figure*}[tb]
    \centering
    \includegraphics[width=0.75\linewidth]{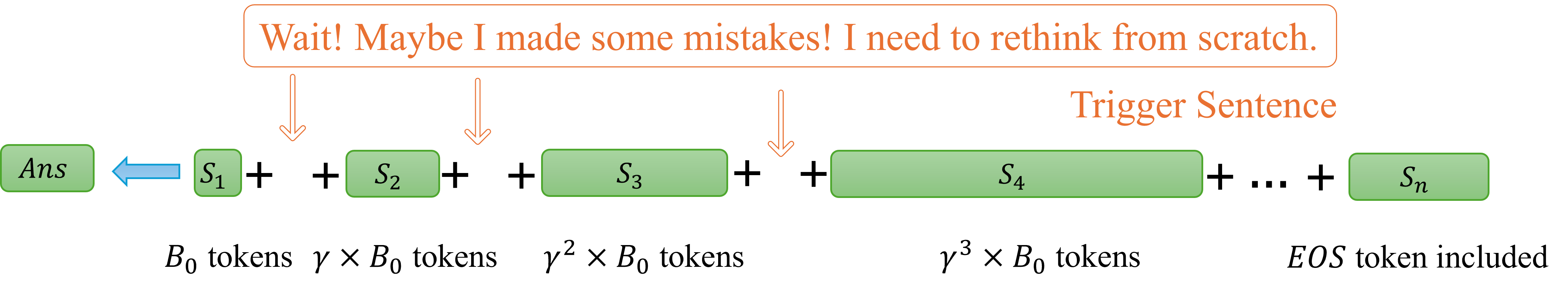}
    \caption{An illustration of our method, Iterative Deepening Sampling (ID-sampling), where $B_0$ is the initial sampling budget.}
    \label{fig:workflow}
\end{figure*}

A key insight in this line of research is that high-quality data capable of eliciting self-reflection in LLMs is crucial for enhancing their reasoning capabilities. Notably, o1-like models demonstrate improved performance when longer reasoning chains are utilized \cite{huang2024o1, deepseekai2025deepseekr1incentivizingreasoningcapability}. One straightforward yet effective approach to collecting such data involves manually inserting prompts such as "Wait! Maybe I made some mistakes! I need to rethink from scratch." These interventions have been shown to successfully trigger self-reflection, thereby improving the model’s reasoning depth and accuracy. However, as this remains an emerging area, many design choices in implementing self-reflection mechanisms lack systematic analysis. In particular, fundamental questions, such as when and how self-reflection should be introduced, remain largely unexplored.

In this paper, we address the challenge of efficient test-time scaling using a fixed model, without additional training, on complex reasoning tasks such as mathematical problem-solving. Specifically, we focus on manually introducing self-reflection triggers during the generation process to improve the pass rate of a fixed model.
To achieve this, we propose Iterative Deepening Sampling (ID-Sampling), a novel algorithmic framework that iteratively increases the sampling budget following a geometric progression, while incorporating self-reflection mechanisms at each expansion step. We theoretically demonstrate that ID-Sampling effectively balances computational efficiency and response quality, ensuring that the budget is not excessively wasted while still improving model performance.
We evaluate ID-Sampling on two challenging reasoning benchmarks, MATH-500 and AIME-24, and demonstrate its effectiveness in Best-of-N sampling and majority voting settings. Additionally, we provide an ablation study analyzing how the rate of budget increase per iteration impacts both pass rate and inference time. Our results highlight the potential of ID-Sampling as a scalable approach for improving LLM reasoning performance through adaptive self-reflection mechanisms.

\section{Related Works}

There are two primary directions for test-time scaling, namely scaling on multiple responses, and scaling the reasoning steps in a single response. 

To efficiently scale to multiple responses, researchers have proposed both aggregating independent samples and employing structured methods such as tree-search-based architectures. While there have been many works on tree-search strategies early on \cite{zhang2023planning, liu2023making, hao2023reasoning, chen-etal-2024-step, zhou2023language, zhang2024accessing, zhang2024llama}, researchers are putting more attention on aggregating independent responses given that it is hard to create a good process reward model to accurately estimate the quality of a partial solution~\cite{deepseekai2025deepseekr1incentivizingreasoningcapability}. Specifically in this direction, researchers mostly rely on two main classes of aggregating strategies, namely self-consistency \cite{wang2022self} and Best-of-N~\cite {snell2024scaling}. There are many works that focus on this setting to make it more efficient and effective. For example, Sun et al. \cite{sun2024fast} proposed to use speculative rejection in BoN to reject bad responses through early scores. Chen et al. \cite{chen2024flaming} proposed to use extremely high temperature on the first token to greatly improve the Best-of-N performance on math and coding tasks. 

To scale within a single response, researchers have first introduced an intermediate step by the so-called chain-of-thoughts\cite{wang2024chain}. Then, starting with Self-Refine \cite{madaan2024self}, there are many works that studied how to effectively use an LLM to give themselves their own evaluation and reflection \cite{yao2022react, shinn2023reflexion}, and study how this reflection-based mechanism can be adapted to different applications \cite{chen2023teaching, gou2023critic, chen2024reprompt}. While there have previously been a lot of discussions on whether LLM can actually self-evaluate and self-correct themselves \cite{huang2023large, chen2024tree, verma2024brittle}, recent studies have shown that by training with high-quality data that involve such a process, LLM can actually achieve a useable level of capability and help self-correct in the process \cite{huang2024o1, zeng2024scaling}. Inspired by the recent success of Deepseek-R1 \cite{deepseekai2025deepseekr1incentivizingreasoningcapability}, people have realized that scaling within a single-response can be efficiently achieved through reinforcement learning, and many follow-up works have also studied how to learn such scaling capability more efficiently \cite{wang2025think}. In this paper, we manually inject \textit{trigger sentence} in the middle of generation, which is similar to the recent work s1 \cite{muennighoff2025s1}. However, our work focuses on injecting at an early stage in the thinking process, while they focus on injecting when the current response completely finishes as a method for budget forcing. 

While the research in the two search strategies is mostly separate, their methods are orthogonal and can be used together to make the sampling process more efficient \cite{snell2024scaling}, and some recent research has also studied how to balance the two, and has shown that scaling on the number of responses can be more efficient than scaling within a single response \cite{wang2025think, marjanovic2025deepseek}. However, their approach is empirically summarized based on a specific group of models distilled from Deepseek-R1.
In this paper, we focus on the intersection of both scaling strategies and study how to strategically trigger the self-correction capability of LLMs more efficiently, considering scaling the number of responses as an option. We take into account the additional cost of scaling within one response by calculating the equivalent $N$ when comparing the answers.

\section{Preliminaries}

\subsection{Best-of-N Sampling}

For reasoning-intensive tasks such as mathematical problem-solving and coding, Best-of-N sampling is one of the most widely used strategies for data generation. This approach involves sampling N outputs from the same model using predefined sampling parameters — typically with a higher temperature than single-sample settings  —  followed by a selection process to determine the best response. The selection criteria depend on the intended use of the samples. During training, responses are typically evaluated using rule-based checkers for mathematical problems or online judges for coding tasks to identify correct answers within the sampled set. At test time, a reward model is often employed to score the generated responses, with the highest-scoring output selected as the final answer. This methodology effectively balances exploration and optimization, making it a fundamental component in enhancing the performance of LLMs on reasoning tasks.

The BoN sampling is a simple yet effective method that can be fully parallelized to enhance performance. Increasing N guarantees improved results during training when a ground-truth checker is available and generally leads to better performance at inference time, provided that the reward model is sufficiently accurate. 
However, if paired with a reward model that is not good enough, BoN might fail to scale efficiently and might even decrease its performance when more samples are included for aggregation. Therefore, getting a good reward model is necessary for good performance for BoN.

\subsection{Majority Voting}

Similar to BoN sampling, majority voting, also known as self-consistency (cons@n), provides an alternative approach for aggregating N different responses \cite{wang2022self}. As the name suggests, this method involves generating N responses, counting the frequency of each unique answer, and selecting the most frequently occurring response as the final output. This approach leverages the inherent redundancy in multiple generations to improve robustness and reliability, making it particularly useful for tasks requiring high confidence in correctness.

Majority voting offers the advantage of aggregating responses efficiently without relying on a reward model. Additionally, it can be extended to a weighted version, where weights are assigned based on PRM scores or confidence estimations of the generated answers \cite{wang2022self}. While majority voting benefits from not requiring a highly accurate reward model, it faces challenges in identifying equivalent answers in complex reasoning tasks. 
For instance, it can be very hard to identify two code samples to be the same, and even in mathematical problems, expressions such as $\frac{1}{\sqrt{3}} $ and $\frac{\sqrt{3}}{3}$ are equivalent but must be recognized as such to ensure correct vote counting. A common solution in mathematical domains involves using symbolic-based checkers to compare answer pairs and identify equivalences. However, this process can be computationally expensive, requiring up to $O(N^2)$ comparisons. In some cases, this comparison overhead can be as time-consuming as the GPU inference itself, posing a significant bottleneck in large-scale reasoning tasks.

\section{Iterative Deepening Sampling}

\subsection{Motivation: Frequencies of Linguistic Markers in Reasoning Models}
\label{sec:motivation}

\begin{figure}[tb]
    \centering
    \includegraphics[width=0.8\linewidth]{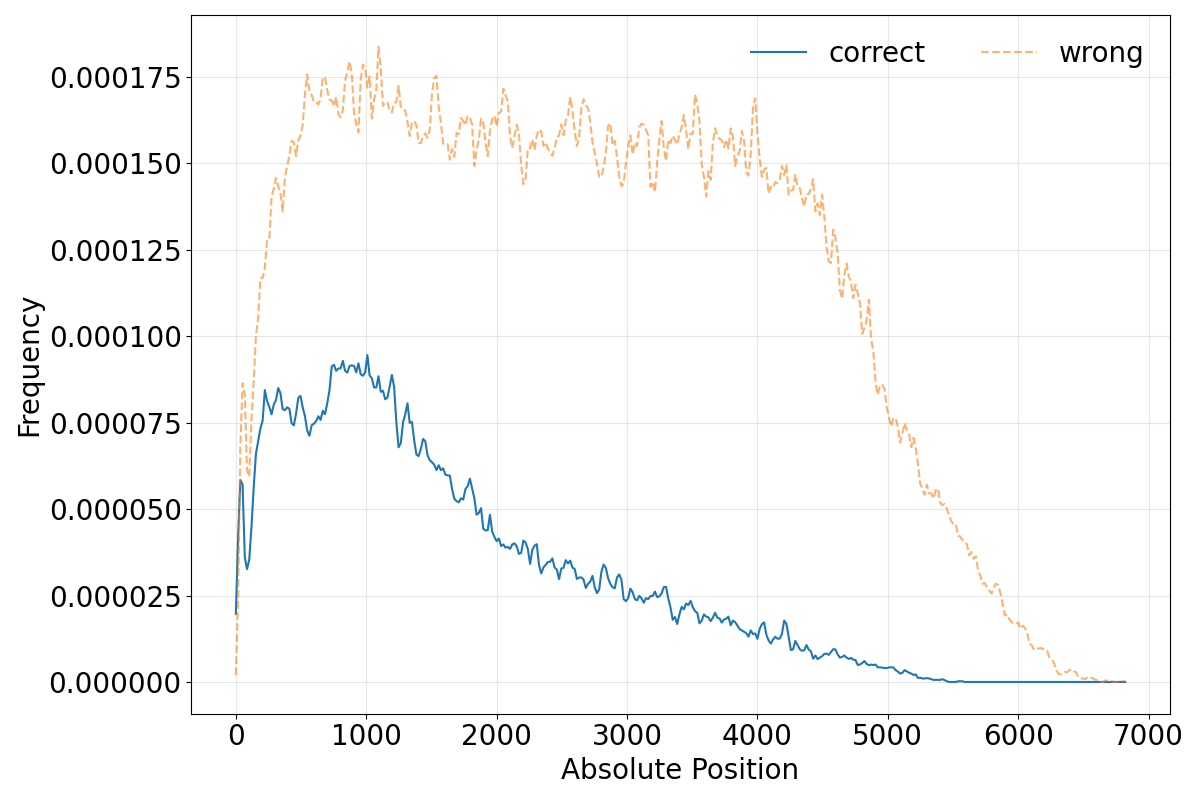}
    \caption{The frequency of linguistic markers related to thinking appeared in Deepseek-R1 on AIME-2024 dataset.}
    \label{fig:marker_frequency}
\end{figure}

Recent studies have highlighted that overthinking can negatively impact the performance of reasoning models \cite{wang2025think, marjanovic2025deepseek}. A key observation is that correct answers tend to be shorter on average and contain fewer linguistic markers associated with "thinking" behaviors.

Here, we present a novel perspective by analyzing not only the frequency but also the positional distribution of thinking-related markers within model responses. A complete list of these markers is provided in the appendix. As illustrated in Fig.~\ref{fig:marker_frequency}, correct answers tend to exhibit fewer think markers, with a clear concentration in the early stages of the response. Their frequency drops sharply after the first 1000 tokens. In contrast, incorrect answers often contain self-corrections distributed throughout the response, with a noticeable decline only near the end. This distinct distribution pattern suggests that marker positioning may serve as a useful signal for assessing answer correctness.

\subsection{Methods}
\begin{algorithm}[thb]
\caption{Iterative Deepening Sampling}
\label{alg:id_sampling}
\begin{algorithmic}[1]
\STATE \textbf{function} \textsc{Generate\_with\_ID-Sampling}(LLM, Dataset, $B_0$, $\gamma$, $B$):
\STATE $Prefixes \gets \textit{Questions from Dataset}$
\STATE $Budget \gets B_0$
\WHILE{$Prefixes$ is not empty \AND $Budget \le B$}
    \STATE $NewPrefixes \gets \{\}$
    \STATE $Outputs \gets \text{LLM.generate}(input=Prefixes, max\_tokens=Budget)$
    \FOR{each $Output$ in $Outputs$}
        \IF{$Output$ is finished}
            \STATE \textsc{LogAnswer}($Output$)
        \ELSIF{$Output$ finished thinking}
            \STATE $Output \gets \text{LLM.generate}(input=Output)$
            \STATE \textsc{LogAnswer}($Output$)
        \ELSE
            \STATE $NewPrefix \gets \textsc{PadTriggerSentence}(Output)$
            \STATE $NewPrefixes.\text{append}(NewPrefix)$
        \ENDIF
    \ENDFOR
    \STATE $Budget \gets Budget \times \gamma$
    \STATE $Prefixes \gets NewPrefixes$
\ENDWHILE
\STATE \textbf{end function}
\end{algorithmic}
\end{algorithm}

In this paper, we focus on efficiently scaling test-time computation by replacing the standard sampling algorithm with a more effective, custom-designed alternative.
Given a fixed overall computational budget, an important challenge is determining how much additional budget should be allocated to refining a given prefix $x_0$ that the model has already sampled. Efficient budget allocation is crucial, as any saved resources can be redirected to increasing the number of N in Best-of-N sampling or deepening tree search, ultimately improving overall performance. Understanding this trade-off is key to optimizing both search efficiency and model output quality.

More specifically, the self-evaluation and self-correction process of the LLM can be manually triggered by introducing a predefined \textit{trigger sentence}, such as "Wait! Maybe I made some mistakes! I need to rethink from scratch." or simply "Wait". In most cases, LLMs respond to this trigger by restarting their reasoning process and making self-corrections. Repeatedly inserting this sentence increases the overall length of the reasoning trajectory, potentially improving problem-solving accuracy by facilitating iterative refinement. In this paper, we propose a method for strategically placing a fixed \textit{trigger sentence} at increasing intervals within longer contextual windows, aiming to balance computational efficiency and the effectiveness of self-correction.

Motivated by the analysis in Section~\ref{sec:motivation}, we propose to introduce more \textit{trigger sentences} at earlier stages and gradually reduce their occurrence as the response length increases.
Suppose we have already used a budget of $b$ to generate a prefix $x_0$ and are now considering whether to immediately introduce a \textit{trigger sentence}. Iterative Deepening (ID) sampling allocates an additional budget of $\gamma \times b$ before inserting the \textit{trigger sentence} the next time, where $\gamma > 1$ is a tunable hyperparameter. This iterative process continues until reaching a maximum budget $B$, beyond which no further \textit{trigger sentences} are introduced.
If the reasoning process reaches a natural stopping point—i.e., a complete answer is generated within the allocated budget—the process terminates early. This is because generating a full response from scratch generally leads to more reliable outputs than attempting to refine an already complete solution. 
The complete procedure is outlined in Algorithm~\ref{alg:id_sampling}, where $B_0$ represents the initial budget. The function LLM.generate conducts the generation within a given budget and is adaptable to different tree-search strategies. The function PadTriggerSentence handles the insertion of \textit{trigger sentences} while ensuring redundancy is minimized if necessary.

The definition and allocation of computational budget depend on the specific test-time scaling algorithm employed, leading to variations in implementation strategies. In this paper, we focus on the setting where N responses are sampled independently, and we have provided an illustration of ID-sampling in Fig.~\ref{fig:workflow}. 
Since response generation is independent and typically performed in parallel, the computational cost primarily depends on the total number of generated responses $N$ and their respective lengths. To manage computational efficiency, we define the budget as the maximum number of tokens generated in a given round, which corresponds to the $max\_token$ parameter in LLM serving engines such as vLLM \cite{kwon2023efficient}. Additionally, to avoid inserting the \textit{trigger sentence} mid-sentence, we extend generation until the completion of a reasoning step. Here, a step is identified by token splits such as `\textbackslash n' or `\textbackslash n\textbackslash n'. This ensures that \textit{trigger sentence} placement aligns with the logical structure of the generated response, preserving coherence and stability in the reasoning process.

Leveraging modern serving engines to efficiently allocate the KV-cache and execute inference in batches using data parallelism and pipeline parallelism significantly enhances the efficiency of ID-sampling. By setting a maximum generation token limit in the sampling parameters, the computational cost still scales linearly with generation time, ensuring that ID-sampling remains efficient even as the number of iterations increases.

\subsection{Theoretical Analysis}

A key challenge in ID-sampling is that budget control occurs before each manually triggered self-evaluation and self-correction step without explicitly analyzing the actual generated responses. This can lead to unnecessary iterations, potentially increasing computational costs. However, due to the design of our ID-sampling method, we establish important theoretical guarantees. In particular, we provide a bound on the total number of tokens generated before reaching the final answer, as formalized in the following theorem.

\begin{theorem}
\label{thm:length}
    Suppose the final answer obtained through ID-Sampling needs a budget of $L$ in normal sampling without manual injection. Then the total number of budget used is no more than $\frac{\gamma * L}{\gamma - 1}$.
\end{theorem}

\paragraph{Proof Sketch}
Note that the budget for each generation iteration follows a geometric sequence with common ratio $\gamma$. The theorem follows directly from a summation of this geometric series.

The theorem does not guarantee the quality of the generated answer or the generation distribution to be the same; rather, it ensures that our method introduces no significant additional computational overhead during the generation of a single response. And we will later show that this overhead will be even smaller in practice. The observed superiority of our algorithm stems from the intuition that explicitly injecting a \textit{trigger sentence} biases the response distribution toward higher-quality outputs.

It is important to note that Iterative Deepening (ID) sampling does not impose any assumptions or constraints on the model’s inherent self-correction or self-evaluation capabilities. In some cases, the model may naturally generate a response that already includes a \textit{trigger sentence} before an explicit manual insertion. We observe that the built-in reasoning capabilities of recent state-of-the-art models, such as OpenAI-o1 \cite{jaech2024openai} and DeepSeek-R1 \cite{deepseekai2025deepseekr1incentivizingreasoningcapability}, significantly impact the effectiveness of ID-sampling. To better understand these effects, we conduct a comprehensive study using DeepSeek-R1, which we present in the experimental section.

\section{Experiments}

\subsection{Experiment Setup}

\label{sec:exp_set}

In our experiments, we evaluate ID-sampling on a set of mathematical problems. Given the strong problem-solving capabilities of modern models, which inherently possess self-evaluation and self-correction mechanisms, we focus on more challenging benchmarks and omit relatively easier datasets like GSM8K \cite{cobbe2021gsm8k}. Instead, we directly assess performance on competition-level and Olympiad-level benchmarks, which are more commonly included in technical reports of recently released models, including MATH-500 \cite{lightman2023lets}, AIME-24 and AIME-25 \cite{MAA_AIME_2025}. Specifically, MATH-500 is a curated subset of 500 problems from the original MATH dataset. Meanwhile, AIME (American Invitational Mathematics Examination) is a highly competitive exam aimed at identifying top-performing high school students in the U.S. The AIME-24 and AIME-25 dataset each comprises 30 problems sourced from the AIME I and II exams, providing a rigorous benchmark for assessing reasoning capabilities under competition-level constraints. 
For both datasets, answers are evaluated using symbolic-based checkers to ensure that all mathematically equivalent responses are recognized as correct.

For our experiments, we use Llama-3.1-1B-Instruct \cite{dubey2024llama} and Phi-4 \cite{abdin2024phi} as examples for non-reasoning models, and DeepSeek-R1-Distill-Qwen-7B, DeepSeek-R1-Distill-Qwen-32B\cite{deepseekai2025deepseekr1incentivizingreasoningcapability} and Qwen3-8B for reasoning models. 
For reward models for BoN, we employ Qwen-2.5-Math-PRM-7B \cite{prmlessons}, which demonstrates superior accuracy compared to other available models. This choice ensures a robust evaluation of reasoning performance in our study.

For baselines, we compare our approach against vanilla sampling without any manual injection. For evaluation, we report three key pass rate metrics across the datasets:
\begin{enumerate}
    \item Best-of-N (BoN) – The accuracy when a reward model selects the best response from N generated samples.
    \item Pass@N - The pass rate, which measures whether or not at least one of the N total responses is correct. We use an unbiased estimation, following the calculation in \cite{chen2021evaluating}.
    \item Majority Voting (cons@N) – The accuracy when responses are aggregated via unweighted majority voting, selecting the most frequent answer.
\end{enumerate}
For both BoN and majority voting, a single aggregated answer is compared against the ground-truth solution to measure accuracy. We report BoN on Math500 datasets and for non-reasoning models on AIME datasets, given that its response length is within the context length limit of the reward models, and report Pass@n for reasoning models on AIME datasets because the normal response length is >12K and beyond the context length of popular reward models. \footnote{We have tried to use techniques like Yarn to extend the context length limit to a longer context. However, the model will lead to a consistent decrease rather than an increase when we increase the number of responses, and thus, we decided to just report Pass@n on longer context answers.}

For \textit{trigger sentence}, we choose to use a whole sentence "Wait! Maybe I made some mistakes! I need to rethink from scratch." for non-reasoning models and a single word "wait" for reasoning models. We will provide a discussion about this choice later. By default, we use $\gamma=2.0$ for ID-sampling, and we will later provide ablation study results on this. Due to the page limit, we leave other hyperparameters in the appendix.

\subsection{Results}

\subsubsection{Math-500}

\begin{figure*}[tb]             %
  \centering

  \begin{subfigure}[t]{0.32\linewidth}
    \centering
    \includegraphics[width=\linewidth]{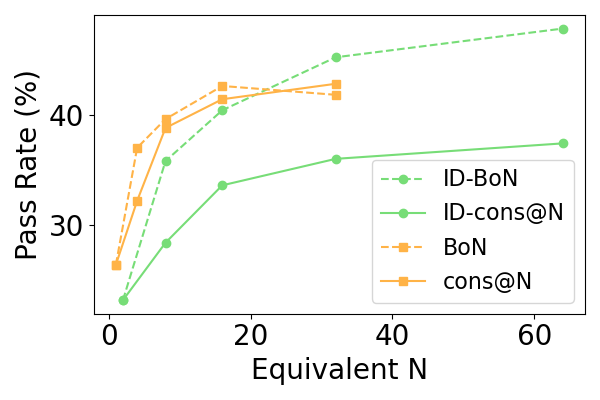}
    \caption{Llama-3.1-1B-Instruct}
    \label{fig:llama_math}
  \end{subfigure}\hfill         %
  \begin{subfigure}[t]{0.32\linewidth}
    \centering
    \includegraphics[width=\linewidth]{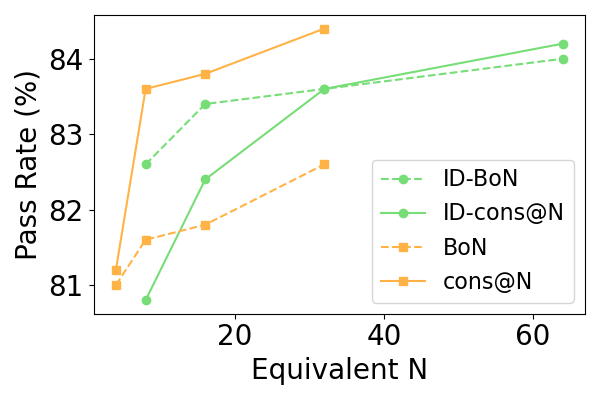}
    \caption{Phi-4}
    \label{fig:phi4_math}
  \end{subfigure} \hfill
  \begin{subfigure}[t]{0.32\linewidth}
    \centering
   \includegraphics[width=\linewidth]{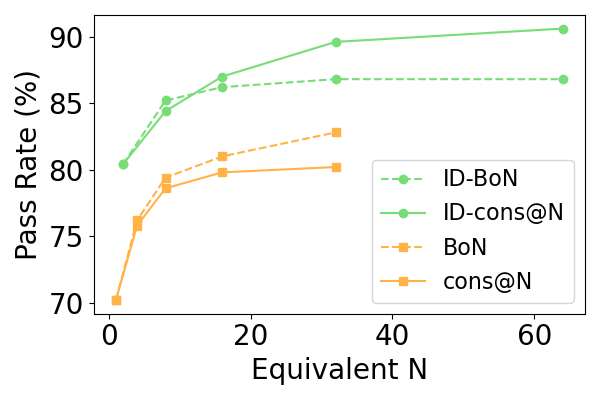}
    \caption{Deepseek-R1-Distill-Qwen-7B}
    \label{fig:r1_math}
  \end{subfigure}
  \caption{Math-500 dataset: Pass rate results for different models. The x-axis is the equivalent N after considering the extra time used by ID-sampling.}
\end{figure*}

Before evaluating the pass rate, we first analyze the runtime overhead of ID-sampling on Math-500 datasets. We observe that ID-sampling incurs approximately 1.6–1.9× the total wall-clock time compared to the baseline of vanilla sampling for non-reasoning models, and 1.1–1.3× for reasoning models. Given that this additional computational cost could instead be allocated to generating more responses and selecting the best one, we present our results in terms of an equivalent N. Specifically, if the original results correspond to Bo8, we report them as equivalent $N=16$, as ID-sampling consistently completes within twice the runtime of the original method.

The MATH-500 dataset is a curated subset of the MATH dataset. Originally used by OpenAI to train process reward models (PRMs) \cite{lightman2023lets}, this dataset benefits from the high accuracy of recently released reward models like Qwen-2.5-Math-PRM \cite{prmlessons}. As a result, in this setting, Best-of-N (BoN) sampling can serve as a reliable sampling method even in the absence of a ground-truth checker.
We present our results for different models on Math-500 in  Fig. \ref{fig:llama_math}, \ref{fig:phi4_math}, \ref{fig:r1_math}. 
We observe a notable difference in the efficacy of ID-sampling for non-reasoning models based on the aggregation method. While ID-sampling yields performance gains with increasing sample size (N) under a Best-of-N (BoN) selection strategy, it consistently performs poorly when results are combined via majority voting. 
In the meantime, ID-sampling stably outperforms regular sampling even when equivalent N is considered. 
Our results also indicate that the performance gap between ID-sampling and vanilla sampling increases as model capabilities improve. Notably, even with the DeepSeek-R1-Distill-Qwen-7B, a strong reasoning model with built-in self-evaluation and self-correction, ID-sampling consistently outperforms vanilla sampling. This is because while stronger models excel at self-correction, they remain suboptimal at determining when to initiate the self-correction process. Compared to earlier models, each \textit{trigger sentence} has a more pronounced effect, allowing ID-sampling to correct errors that might otherwise persist without explicit intervention.

\subsubsection{AIME}

\begin{table}[tb]
\centering

  \begin{subtable}[t]{0.48\textwidth}
    \centering
    \begin{tabular}{c|cc|cc}
      \hline
      $N$ & \multicolumn{2}{c|}{\textbf{Vanilla}}
          & \multicolumn{2}{c}{\textbf{ID-sampling}}\\\cline{2-5}
          & \textbf{BoN} & \textbf{cons@N} & \textbf{BoN} & \textbf{cons@N}\\\hline
      1   & 0.00 & 0.00 & \textbf{3.45} & \textbf{3.45}\\
      4   & 0.00 & 0.00 & 0.00 & \textbf{3.45}\\
      8   & 0.00 & 0.00 & 0.00 & \textbf{3.45}\\
      16  & \textbf{3.45} & 0.00 & 0.00 & \textbf{3.45}\\
      32  & 3.45 & 0.00 & \textbf{10.34} & \textbf{6.90}\\\hline
    \end{tabular}
    \caption{Llama-3.1-1B-Instruct}\label{tab:aime_non_r1_llama}
  \end{subtable}
  \hfill
  \begin{subtable}[t]{0.48\textwidth}
    \centering
    \begin{tabular}{c|cc|cc}
      \hline
      $N$ & \multicolumn{2}{c|}{\textbf{Vanilla}}
          & \multicolumn{2}{c}{\textbf{ID-sampling}}\\\cline{2-5}
          & \textbf{BoN} & \textbf{cons@N} & \textbf{BoN} & \textbf{cons@N}\\\hline
      1   & 17.24 & 17.24 & \textbf{20.69} & \textbf{20.69}\\
      4   & 13.79 & 17.24 & \textbf{24.14} & \textbf{24.14}\\
      8   & 13.79 & 20.69 & \textbf{27.59} & \textbf{24.14}\\
      16  & 13.79 & 20.69 & \textbf{27.59} & \textbf{20.69}\\
      32  & 13.79 & 20.69 & \textbf{27.59} & \textbf{20.69}\\\hline
    \end{tabular}
    \caption{Phi-4}\label{tab:aime_non_r1_phi}
  \end{subtable}

\caption{AIME-24 dataset: Pass rate results for different models with different number of samples. 
The best results in each setting are highlighted in bold. Since performance saturates quickly on these weaker LLM models, meaning that even the highest pass rate with $N=32$ does not exceed the $N=1$ pass rate for ID-sampling, we omit the use of equivalent N in this setting.
}
\label{tab:aime_non_r1}
\end{table}

\begin{figure*}[tb]
  \centering

  \begin{minipage}[t]{0.33\textwidth}
    \vspace{0pt}%
    \centering
    \includegraphics[width=\linewidth]{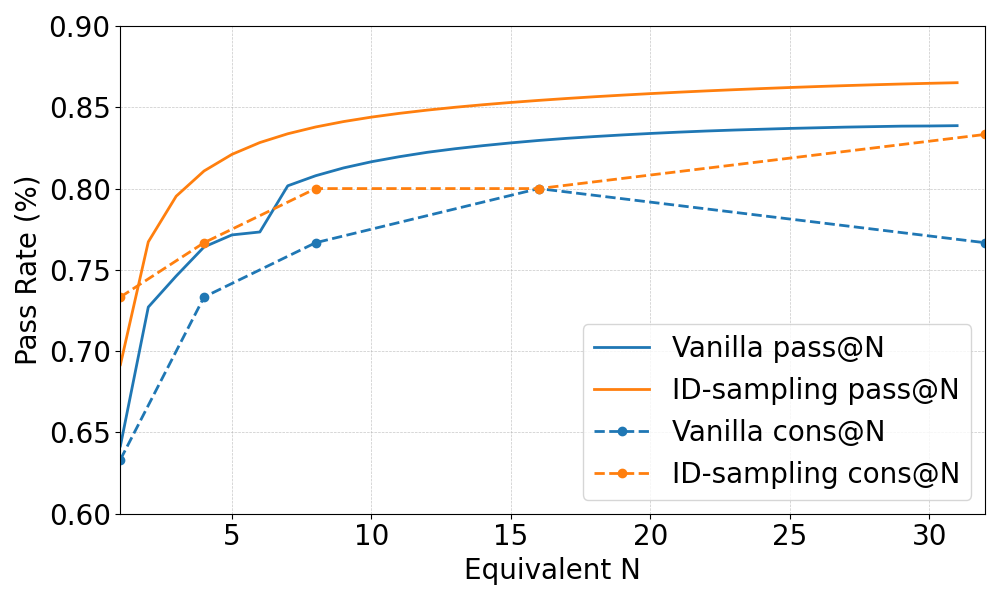}
    \captionof{figure}{Qwen3-8B results on AIME-24. }
    \label{fig:qwen3_aime24}
  \end{minipage}\hfill
  \begin{minipage}[t]{0.33\textwidth}
    \vspace{0pt}
    \centering
    \includegraphics[width=\linewidth]{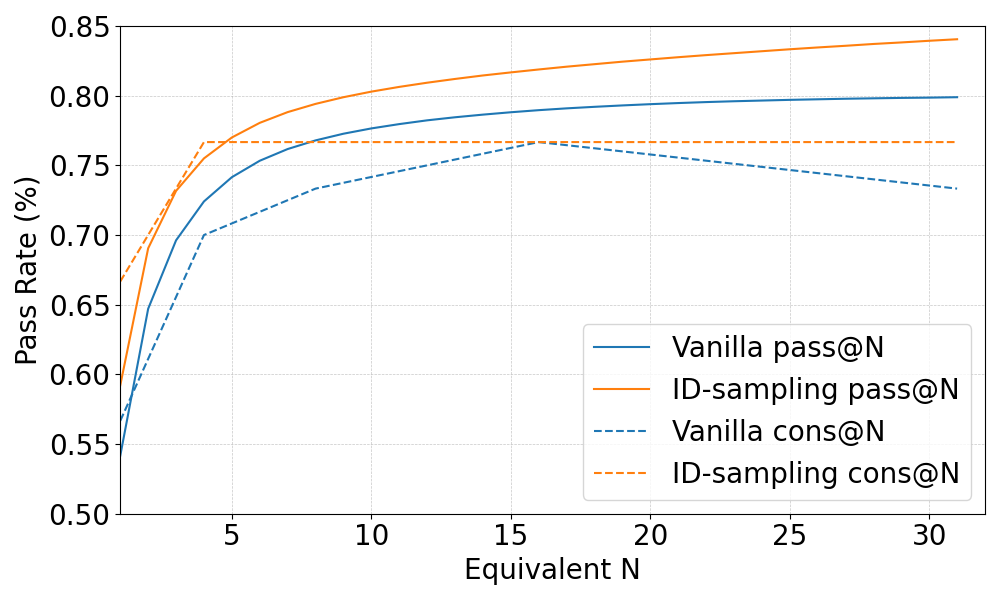}
    \captionof{figure}{Qwen3-8B results on AIME-25.}
    \label{fig:qwen3_aime25}
  \end{minipage}\hfill
  \begin{minipage}[t]{0.33\textwidth}
    \vspace{0pt}
    \centering
    \includegraphics[width=\linewidth]{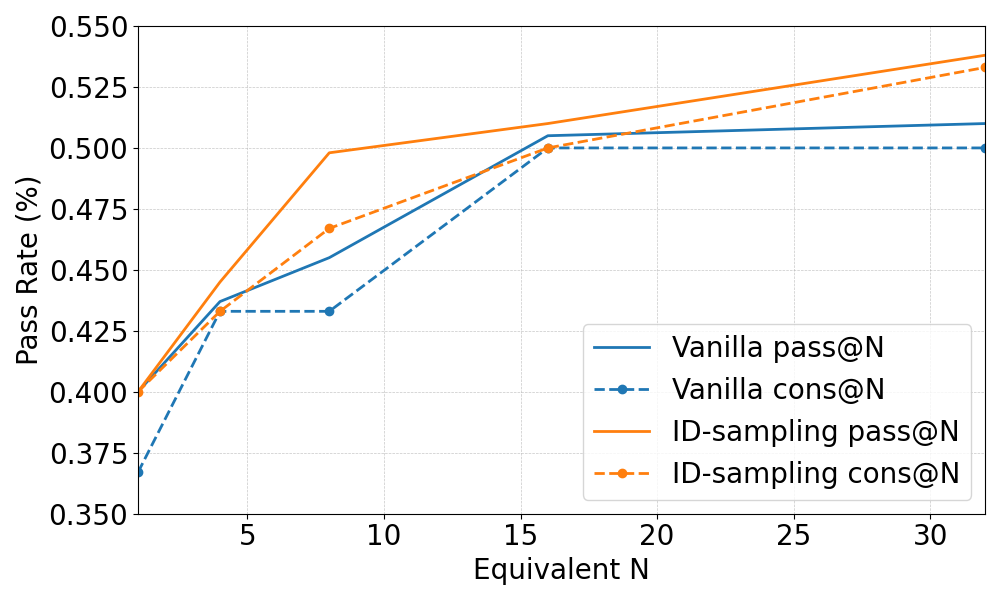}
    \captionof{figure}{Deepseek-R1-distill-Qwen-7B results on AIME-25.}
    \label{fig:r1_aime25}
  \end{minipage}
\end{figure*}

\begin{figure*}[tb]
    \begin{minipage}[t]{0.33\textwidth}
      \vspace{0pt}
      \centering
      \begin{tabular}{lcc}
        \toprule
                 & \textbf{Pass} & \textbf{cons} \\
        \midrule
        Vanilla  & 72.6 & 83.3 \\
        ID-sampling       & 75.5 & 86.7 \\
        \bottomrule
      \end{tabular}
      \captionof{table}{Pass@1 and cons@32 comparison between Vanilla sampling and ID-sampling on Deepseek-R1-distill-Qwen-32B on AIME-24 dataset.}
      \label{tab:pass1_cons}
    \end{minipage}\hfill
    \begin{minipage}[t]{0.65\textwidth}
    \vspace{0pt}
    \centering
    \begin{tabular}{@{}l*{5}{c}@{}}
    \toprule
                & Vanilla & ID(2.5) & ID(2.0) & ID(1.5)  \\ \midrule
    Rel. Time & 1.00   & 1.07   & 1.09   & 1.39    \\
    \bottomrule
    \end{tabular}
    \captionof{table}{Relative wall-clock time cost of ID-sampling for varying $\gamma$
      values using DeepSeek-R1-Distill-Qwen-7B on the AIME-24 dataset.}
    \label{table:time_gamma}
  \end{minipage}
\end{figure*}

\begin{figure*}[t]
  \centering
  \begin{subfigure}[t]{0.45\textwidth}
    \vspace{0pt}%
    \centering
    \includegraphics[width=0.95\linewidth]{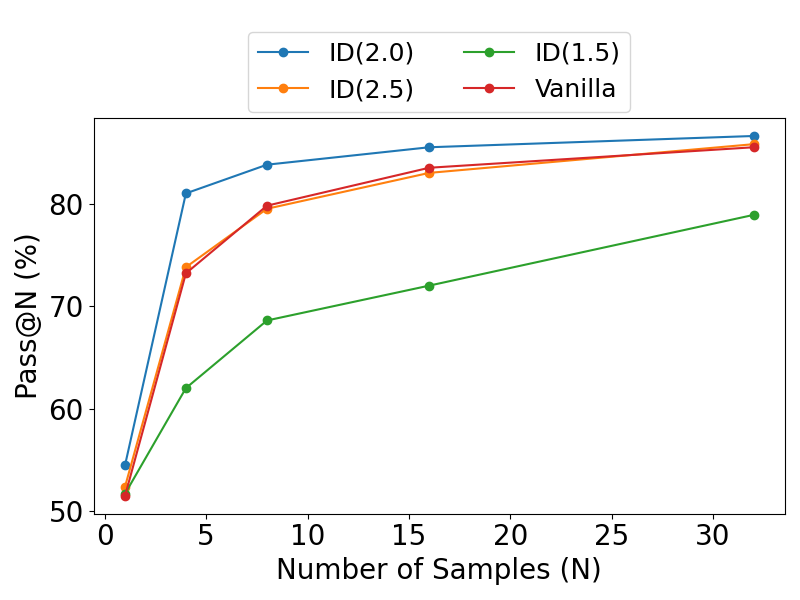}
    \caption{Pass@N}
    \label{fig:r1_bon_aime}
  \end{subfigure}\hfill
  \begin{subfigure}[t]{0.45\textwidth}
    \vspace{0pt}
    \centering
    \includegraphics[width=0.95\linewidth]{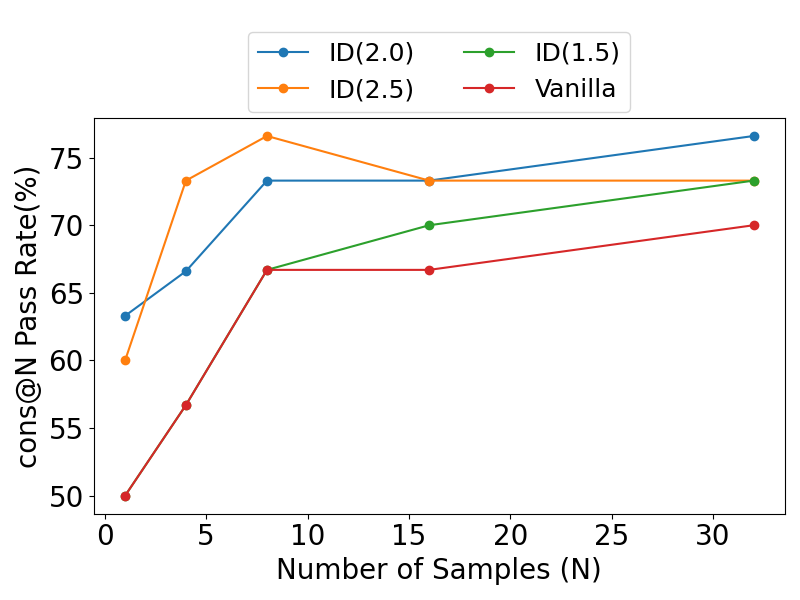}
    \caption{cons@N}
    \label{fig:r1_cons_aime}
  \end{subfigure}
  \caption{AIME-24 Dataset: Pass rate results for DeepSeek-R1-Distill-Qwen-7B with different $\gamma$ for ID-sampling, compared to vanilla sampling. The numbers in brackets are the $\gamma$ used in ID-sampling. As different values of $\gamma$ result in significantly varying runtimes, as shown in Table~\ref{table:time_gamma}, we again omit the use of equivalent $N$ in this analysis.}
  \label{fig:r1_aime24}
\end{figure*}

The AIME datasets, including both AIME-24 and AIME-25, are highly challenging benchmarks that are now widely used as challenging datasets for reasoning models \cite{deepseekai2025deepseekr1incentivizingreasoningcapability, huang2024o1}. In this setting, reward models exhibit significantly lower accuracy, and the reasoning is generally longer.

We first present the results for non-reasoning models in Table~\ref{tab:aime_non_r1}. We observe that ID-sampling can help improve the pass rate, and because of the fast saturation of pass rate, ID-sampling can even help the model to surpass its previous ceiling, i.e., $N=1$ with ID-sampling is better than $N=32$ with vanilla sampling. 

For reasoning models, we first observe that the runtime difference between vanilla sampling and ID-sampling becomes negligible. For Qwen3 models, the time ratio remains within $1 \pm 0.02$, and for R1-distill models, it is within $1 \pm 0.1$ with $\gamma=2.0$. As a result, the use of equivalent $N$ has little practical effect in this context. We present results on reasoning models in Fig.~\ref{fig:qwen3_aime24}, \ref{fig:qwen3_aime25}, \ref{fig:r1_aime25}, \ref{fig:r1_aime24}, and Table~\ref{tab:pass1_cons}. Across different models on the two latest AIME datasets, ID-sampling consistently outperforms vanilla sampling. Specifically, on both datasets, Qwen3-8B achieves a larger improvement margin compared to Deepseek-R1-distill-Qwen-7B. These results highlight not only the robustness of our algorithm on recent models and benchmarks, but also the promise of ID-sampling as models gain stronger built-in self-correction capabilities and stronger reasoning capabilities.

\paragraph{Ablation Study} Given the stable performance gains of ID-sampling on AIME, here we conducted an ablation study to analyze the impact of the scaling factor 
$\gamma$ on ID-sampling. We present the pass rate results for DeepSeek-R1-Distill-Qwen-7B on AIME-24 in Figures \ref{fig:r1_bon_aime} and \ref{fig:r1_cons_aime}, and report the relative inference time for each setting in Table \ref{table:time_gamma}.

We find that adjusting $\gamma$ significantly impacts both performance and computational cost. In terms of runtime, $\gamma = 1.5$ yields the highest cost, whereas the other two settings remain within $1.1\times$ the wall-clock time of vanilla sampling. Regarding performance, $\gamma = 2.0$ consistently achieves the best results in terms of Pass@$N$, while $\gamma = 2.5$ occasionally outperforms in terms of Cons@$N$, though with a small margin. Overall, ID-sampling proves to be a more effective sampling strategy than vanilla sampling, provided that $\gamma$ is not too small, which would undermine our goal of emphasizing early-stage \textit{trigger sentence} injection.

Interestingly, we observed a non-convex relationship between $\gamma$ and performance even without considering the time used, with the second-best setting being $\gamma=2.0$. This non-convex behavior complicates hyperparameter selection. However, as we described in our motivation section, we want our algorithm to inject more in the early stage and drop fast in later stage, and thus recommend using $\gamma=2.0$ given its balance between performance and time.

\section{Discussions}
\label{sec:discuss}

\paragraph{Choice of Trigger Sentence}
In our experiments, we use different \textit{trigger sentences} for non-reasoning models and reasoning models. The difference is caused by the nature of the models. For non-reasoning models, a single word "wait" cannot trigger the self-correction process and can only introduce noise to the generation. For reasoning models, we found that for unknown reason, adding the whole sentence will trigger the generation of an end-of-think token for a certain probability, and leads to early stop in the generation sequence. While this does not always harm the performance, this is not something we want, as we want to use the \textit{trigger sentence} to trigger self-correction. Overall, we believe the choice of \textit{trigger sentence} should be case-by-case, depending on the underlying model and the specific usage of the algorithm. But in principle, a single word, "wait," will be sufficient for most reasoning models.

\paragraph{Limitations}Our proposed method also has several clear limitations. The most significant is that the performance of our method could significantly depend on the underlying model. While we have used our experiments to show that ID-sampling works on popular models, it is possible for some models to be incompatible with the algorithm, especially if the models are trained with losses that incorporate more than just the accuracy of the final answer.
Moreover, incorporating \textit{trigger sentences} into the generation process requires ID-sampling to invoke multiple generation steps. While this theoretically incurs no additional cost on the KV-cache, in practice, it can lead to increased inference time unless one modifies the inference engines like vLLM and SGLang to manually store and reuse the KV-cache. To address this concern, we use equivalent $N$ in our experiments, which accounts for total wall-clock time, to enable a fair comparison across different sampling methods.
Additionally, due to the need for multiple sampling steps, ID-sampling is currently tested only on open-source models. While the same idea can also be applied to black-box models through multi-round generations, this will introduce extra assistant tokens during generation, which may cause a slight distribution shift when applied to black-box models. However, our experiments indicate that the impact of this shift is generally minimal and can be ignored in most cases.

\section{Conclusions}

In this paper, we introduce Iterative Deepening (ID) Sampling, a simple yet effective sampling algorithm designed to more efficiently scaling test-time compute compared to regular sampling. We demonstrate that ID-sampling effectively enhances inference-time performance by improving the pass rate on challenging mathematical reasoning tasks across various reasoning models on math problems. We also found that ID-sampling is less effective on non-reasoning models given their weaker self-correction capability. Our results indicate that while current models exhibit strong self-correction capabilities, they still have room for improvement in determining when to trigger self-correction themselves without human intervention, which might not be sufficiently solved by the current reinforcement learning paradigm.

\bibliography{example_paper}
\bibliographystyle{icml2025}

\appendix

\section{Additional Discussions}

\paragraph{Potential Extension to Search-based Test-time Scaling}
In this paper, our study primarily focuses on ID-sampling without integrating it into tree-search algorithms. However, we emphasize that our proposed framework can be directly extended to methods such as beam search and Monte Carlo Tree Search (MCTS). In these cases, the budget in Algorithm~\ref{alg:id_sampling} can be directly interpreted as the computational budget in MCTS or the number of iterations in beam search.
One challenge in this extension is that self-correction mechanisms reduce the reliance on early model outputs being correct, distinguishing the need for a new generation of PRMs from previously released ones.
However, future researchers who develop or have access to more accurate PRMs that explicitly model the self-correction process can leverage our ID-sampling framework to enhance the reasoning capabilities of fixed models.

\paragraph{Choice of Datasets}

In our experiments, we focus exclusively on mathematical reasoning datasets. While test-time scaling techniques can, in principle, be extended to more general domains, such extensions are not directly applicable in our setting. From a methodological standpoint, the injection of \textit{trigger sentences} is not well suited to code generation tasks. In such cases, the insertion of additional tokens can introduce syntactic or semantic errors. More broadly, this work aims to develop a more efficient test-time scaling strategy. One promising direction is to leverage high-quality reward models to aggregate multiple responses, as demonstrated in our approach. However, existing open-source reward models are predominantly designed for mathematical problem solving, and robust, general-purpose reward models are currently lacking. Additionally, popular benchmarks such as MMLU \cite{wang2024mmlu} consist primarily of single-choice questions. For these tasks with a limited number of options, evaluation metrics like Pass@$n$ (for $n > 1$) are not meaningful, as simply generating diverse final options can trivially improve performance. As a result, most standard non-mathematical datasets are not well aligned with the objectives of this study. We leave the exploration of ID-sampling on less commonly used datasets as an avenue for future research.

\section{Linguistic Markers}

In the paper, we have presented the occurrence frequencies of the linguistic markers as our motivation. We provide the full list of such markers here, which is the same as \cite{wang2025think}: "however", "wait", "alternatively", "hmm".

\end{document}